\documentclass[sigconf,nonacm]{acmart}

\setcopyright{none}
\AtBeginDocument{%
  }

\usepackage{amssymb}
\usepackage{booktabs} 
\usepackage{pifont}   
\usepackage{xcolor}   
\usepackage[table]{xcolor}
\usepackage{colortbl} 
\usepackage{pgfmath}

\usepackage[most]{tcolorbox}

\usepackage{multirow}
\usepackage{makecell}

\usepackage{array}

\usepackage{subcaption}
\usepackage{enumitem}

\begin{document}

\title{HotComment: A Benchmark for Evaluating Popularity of Online Comments}


\author{Yafeng Wu}
\affiliation{%
  \institution{Huazhong University of Science and Technology}
  \city{Wuhan}
  \country{China}
}

\author{Yunyao Zhang}
\affiliation{%
  \institution{Huazhong University of Science and Technology}
  \city{Wuhan}
  \country{China}
}

\author{Liliang Ye}
\affiliation{%
  \institution{Huazhong University of Science and Technology}
  \city{Wuhan}
  \country{China}
}

\author{Guiyi Zeng}
\affiliation{%
 \institution{Huazhong University of Science and Technology}
 \city{Wuhan}
 \country{China}
 }

\author{Junqing Yu}
\affiliation{%
 \institution{Huazhong University of Science and Technology}
 \city{Wuhan}
 \country{China}
 }

\author{Chen Xu}
\affiliation{%
  \institution{Beijing Institute of Computer Technology and Applications}
  \city{Beijing}
  \country{China}
}

\author{Zikai Song}
\authornote{Corresponding author.<skyesong@hust.edu.cn>}
\affiliation{%
  \institution{Huazhong University of Science and Technology}
  \city{Wuhan}
  \country{China}
}

\settopmatter{authorsperrow=4}



\begin{abstract}
Online comments play a crucial role in shaping public sentiment and opinion dynamics on social media.
However, evaluating their popularity remains challenging, not only because it depends on linguistic quality, originality, and emotional resonance, but also because stylistic preferences vary widely across platforms and user groups, causing the same comment to resonate differently in different communities.
%
In this work, we present \textbf{HotComment}, a multimodal benchmark integrating video and text modalities  that comprehensively quantifies popularity from three enhanced aspects: (1) Content Quality, which evaluates semantic similarity with ground-truth human comments and extends quality assessment through four interpretable dimensions; (2) Popularity Prediction, based on trends from models trained on real-world interaction data; and (3) User Behavior Simulation, which models the distribution of platform users and approximates \textbf{engagement scores} through an agent-based framework.
Furthermore, we propose \textbf{StyleCmt}, inspired by \textbf{social ripple effects}, where \textbf{multiple stylistic dimensions align} to amplify socially resonant expressions and suppress incongruent ones.
\end{abstract}




\maketitle

\begin{figure}[t]
    \centering
    \includegraphics[width=\columnwidth]{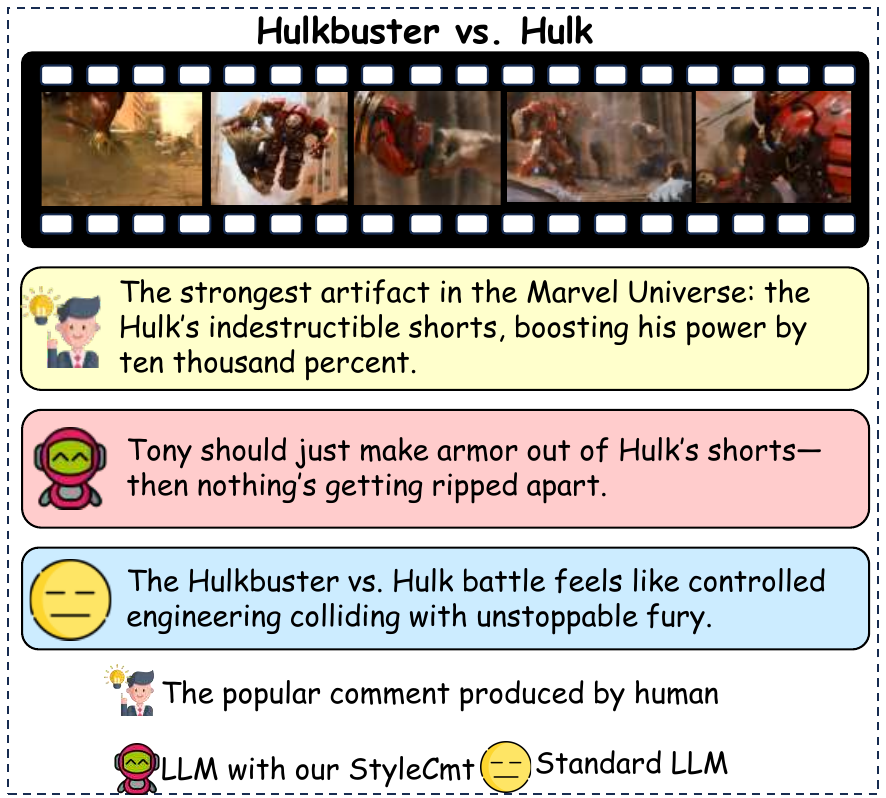}
    \caption{
    \textbf{Example from the HotComment benchmark.} 
    Illustration of three types of comments for a given video: 
    a real \textbf{human-generated popular comment} from the online platform, 
    a \textbf{StyleCmt-enhanced LLM-generated comment}, 
    and a \textbf{standard LLM-generated comment} without stylistic guidance.  
    }
    \label{fig:teaser}
\end{figure}

\section{Introduction}
\label{sec:intro}

Online comments play a central role in shaping discourse on social media platforms\cite{10.1371/journal.pone.0142390, song7, song9}. 
With the rapid advancement of Artificial Intelligence and multimodal technologies~\cite{song5,song6,Retrack, HABIT, OFFSET}, the landscape of popular content generation on social media has shifted from being purely human-driven to increasingly dominated by AI-generated content~\cite{song14, song12}. However, evaluating their popularity remains challenging~\cite{chang2024survey}. Popular comments are not determined by content relevance alone, but also by stylistic expression\cite{lei-etal-2025-godbench}, context-aware wording\cite{10.1145/3664647.3681195}, and the social pathways through which audiences encounter and react to content\cite{berger2012what, bakshy2012role}.

Existing evaluation frameworks for social media comments mainly follow two directions. One line relies on lexical or semantic similarity metrics, such as BLEU and BERTScore\cite{papineni-etal-2002-bleu, zhang2020bertscoreevaluatingtextgeneration}, to measure closeness to human references. Another extends evaluation from stylistic perspectives, incorporating factors such as humor, rhetorical devices, and creativity to better approximate human preference\cite{10.1145/3539597.3570431, Chen_Yuan_Liu_Liu_Guan_Guo_Peng_Liu_Li_Xiao_2024, Zhong_2024_CVPR}. Recent benchmark efforts further emphasize multidimensional stylistic quality in comment-related generation settings\cite{lei-etal-2025-godbench}.
While these methods capture important aspects of comment quality, they still provide only a partial account of popularity. Similarity-based metrics mainly reflect textual overlap or semantic closeness, whereas stylistic indicators are often treated as intrinsic properties of the comment itself. In practice, however, socially valued comments also depend on reasoning quality, constructiveness, and audience-sensitive engagement cues\cite{gottipati2012finding, kolhatkar2017constructive, risch2020top, fujita2019dataset}. Moreover, audience preferences vary across user groups, and exposure itself is shaped by demographic and cultural differences\cite{garrett2009echo, lee2017incidental, schaefer2023incidental}. As a result, treating popularity as a universal textual property makes it difficult for existing frameworks to capture real engagement dynamics and the heterogeneous mechanisms underlying comment popularity\cite{weng2012competition, garrett2013turn}.


To address these challenges, we propose \textbf{HotComment}, a benchmark for evaluating the popularity of online comments.
HotComment introduces a novel three-dimensional evaluation framework:
(1) \textbf{Content Quality}, which evaluates semantic similarity with ground-truth human comments and additionally incorporates four stylistic dimensions ([Linguistic Expression], [Creative Imagination], [Emotional Resonance], and [Social and Cultural Influence]) to complement overlap-based metrics and capture how linguistic artistry, creativity, affective depth, and cultural propagation jointly relate to popularity;
(2) \textbf{Popularity Prediction}, based on scores from engagement prediction models trained on large-scale real-world interaction data; and
(3) \textbf{User Behavior Simulation}, which models heterogeneous user preferences through agent-based simulations of \textbf{engagement scores}.
This multi-perspective design offers a more realistic and interpretable assessment of whether generated comments possess true popularity potential, establishing a robust foundation for future research on popularity-aware generation.

Inspired by the \textit{Wave Interference Model}\cite{hecht2016optics} and the \textit{Uses and Gratifications Theory}\cite{katz1973uses}, we propose \textbf{StyleCmt}, a novel framework that models the interaction among stylistic elements in linguistic space based on the principles of constructive and destructive interference.
StyleCmt captures how different expressive patterns combine to amplify socially resonant forms while attenuating less compatible ones, generating comments that align with collective audience preferences.
Experimental results demonstrate that this framework enables models to produce comments that more closely reflect human preference tendencies.

\newcommand{\cmark}{\color{green}\ding{51}}   
\newcommand{\xmark}{\color{red}\ding{55}}   

\begin{table}[t]
\centering
\small
\resizebox{\columnwidth}{!}{%
\begin{tabular}{@{}l@{\hspace{-2pt}} c c c c@{\hspace{-0.01pt}} c@{\hspace{-0.01pt}} c@{\hspace{-0.01pt}} c@{\hspace{-0.01pt}} c@{\hspace{-0.1pt}} c@{}}
\toprule
\multirow{2}{*}{\textbf{Benchmark}} & 
{\textbf{Multi}} &
\multicolumn{2}{c}{\textbf{Scale}} &
\multicolumn{4}{c}{\textbf{Content Quality}} &
\textbf{Cross} & 
\textbf{Aud.} \\
\cmidrule(lr){3-4}\cmidrule(lr){5-8}
 & \textbf{-modal} & \textbf{Vis.} & \textbf{Txt.} & \textbf{LE} & \textbf{CI} & \textbf{ER} & \textbf{SCI} & \textbf{-plat.} & \textbf{Var.} \\
\midrule
TalkFunny\cite{Chen_Yuan_Liu_Liu_Guan_Guo_Peng_Liu_Li_Xiao_2024} & \xmark & -- & 4k & \cmark & \xmark & \cmark & \xmark & \xmark & \xmark \\
Chumor 2.0\cite{he-etal-2025-chumor} & \xmark & -- & 3k & \cmark & \xmark & \cmark & \xmark & \xmark & \xmark \\
Puns\cite{xu-etal-2024-good} & \xmark & -- & 2k & \cmark & \cmark & \xmark & \xmark & \xmark & \xmark \\
Oogiri-GO\cite{Zhong_2024_CVPR} & \cmark & 100k & 30k & \cmark & \cmark & \xmark & \xmark & \xmark & \xmark \\
NYT-Captions\cite{hessel-etal-2023-androids} & \xmark & 3k & -- & \xmark & \xmark & \xmark & \xmark & \xmark & \xmark \\
ViCo\cite{10.1145/3696409.3700260} & \cmark & 20k & -- & \cmark & \xmark & \cmark & \xmark & \xmark & \xmark \\
HOTVCOM\cite{chen-etal-2024-hotvcom} & \cmark & 93k & -- & \cmark & \cmark & \cmark & \cmark & \xmark & \xmark \\
GODBench\cite{lei-etal-2025-godbench} & \cmark & 67k & -- & \cmark & \cmark & \cmark & \cmark & \xmark & \xmark \\
\midrule
\textbf{HotComment} & \cmark & 34k & 47k & \cmark & \cmark & \cmark & \cmark & \cmark & \cmark \\
\bottomrule
\end{tabular}
} 
\caption{
\textbf{Benchmark comparison.}
\textbf{Vis.} and \textbf{Txt.} denote \emph{Visual} and \emph{Text}.
\textbf{LE}, \textbf{CI}, \textbf{ER}, and \textbf{SCI} denote the four stylistic dimensions of \emph{Content Quality}.  
\textbf{Cross-plat.} indicates datasets supporting cross-platform evaluation,  
and \textbf{Aud. Var.} denotes those considering audience variation.
}
\label{tab:benchmark_comparison}
\end{table}

Our contributions are summarized as follows:
\begin{itemize}[leftmargin=*]
    \item We introduce a new large-scale and comprehensive online comment dataset with a multidimensional evaluation framework for online comments tasks.
    \item We propose StyleCmt, which simulates the interactions among different stylistic patterns, thus enabling the model to produce comments that resonate with dominant user preferences.
    \item Extensive experiments on the HotComment benchmark demonstrate that StyleCmt effectively captures stylistic and social dynamics, significantly improving the realism and engagement alignment of generated comments compared with baselines.
\end{itemize}


\section{Related Work}
\label{sec:related}

\subsection{Evaluation of comment quality}
Early research on comment evaluation mainly focused on intrinsic textual quality, emphasizing coherence, fluency, and semantic consistency~\cite{HINT,REFINE,INTENT} with human-written references. Lexical and semantic metrics such as BLEU~\cite{papineni-etal-2002-bleu}, ROUGE~\cite{lin-2004-rouge}, and BERTScore~\cite{zhang2020bertscoreevaluatingtextgeneration} have been widely adopted to measure textual overlap and meaning alignment.
Subsequent studies extended this line of work by incorporating stylistic and rhetorical dimensions into comment assessment. Existing research has examined humor~\cite{10.1145/3539597.3570431, Chen_Yuan_Liu_Liu_Guan_Guo_Peng_Liu_Li_Xiao_2024}, irony~\cite{lin2024augmentingemotionfeaturesirony}, creativity~\cite{Zhong_2024_CVPR}, puns~\cite{sun2022expunationsaugmentingpunskeywords, xu-etal-2024-good}, emotional expressiveness, and broader multidimensional stylistic quality~\cite{lei-etal-2025-godbench}. These efforts deepen the understanding of how rhetorical and expressive features contribute to perceived comment quality.
Beyond intrinsic quality and stylistic sophistication, another work has studied thoughtful and constructive comments as a distinct form of socially valued response. These studies show that constructiveness, reasoning quality, and conversational usefulness are not always equivalent to raw popularity feedback such as likes or replies, and should often be modeled separately~\cite{gottipati2012finding, kolhatkar2017constructive, fujita2019dataset, kobayashi2021case, risch2020top}.
Studies on constructive comments show that constructiveness, reasoning quality, and conversational usefulness often diverge from raw popularity signals like likes or replies, and should be modeled separately~\cite{gottipati2012finding, kolhatkar2017constructive, fujita2019dataset, kobayashi2021case, risch2020top}. A similar mismatch occurs in image retrieval, where superficial engagement cues do not reliably reflect true relevance or user utility~\cite{Air-Know, ConeSep, TEMA, ENCODER}. These converging findings suggest that valuable feedback must often be disentangled from coarse popularity metrics.

Overall, existing comment evaluation methods mainly focus on intrinsic textual and stylistic quality, providing limited support for modeling platform-dependent engagement patterns and heterogeneous audience preferences. To address this gap, \textbf{HotComment} extends evaluation beyond the comment itself by introducing two complementary dimensions: \textbf{popularity prediction} and \textbf{user behavior simulation}.

\subsection{Comment generation}
Early studies on automatic comment generation~\cite{zheng2017gated, qin2018automatic, ma2018unsupervised} mainly relied on deep learning frameworks based on attention mechanisms~\cite{song11, STABLE}, structured modeling~\cite{song8,song10,song13}, or unsupervised matching between articles and comments. Later approaches improved diversity and structural representation through graph-based and retrieval-augmented designs~\cite{li2019graph, yang2019read, yang2019cross}, but they remained limited in semantic depth and rhetorical control.
With the rise of large language models~\cite{chang2026decomposing,chang2026balora,chang2025lora,song15}, comment generation has advanced substantially in multi-modal settings~\cite{10.1145/3696409.3700260, chen-etal-2024-hotvcom, li2025miv, li2025taco}, include both both text and visual cues~\cite{song1,song2,song3,song4}. Recent studies have explored personalized comment generation~\cite{zeng2019automatic, lin2024personalized}, popularity-aware social response generation~\cite{yu2024popalm}, and controllable style steering during inference~\cite{zhang2025personalized}, showing that user identity, stylistic preference, and anticipated audience response are important conditioning factors. Existing methods, however, typically improve only a specific aspect of generation, such as creativity~\cite{Zhong_2024_CVPR}, humor~\cite{Chen_Yuan_Liu_Liu_Guan_Guo_Peng_Liu_Li_Xiao_2024}, or contextual relevance~\cite{10.1145/3696409.3700260}. In the multimodal setting, increasingly realistic tasks and datasets, such as LiveBot~\cite{ma2019livebot}, PLVCG~\cite{zeng2021plvcg}, knowledge-enhanced live video comment generation~\cite{chen2023knowledge}, and MMLSCU~\cite{meng2024mmlscu}, have further enriched the problem setting. Nevertheless, most existing methods still treat controllable factors such as style, humor, or preference as isolated attributes, rather than explicitly modeling how multiple stylistic tendencies interact within a comment community and jointly shape audience resonance. To address this limitation, we propose \textbf{StyleCmt}, a cross-platform and multimodal framework that models comment-section preferences and generates comments aligned with dominant audience styles and expectations.

\section{Challenge and Motivation}

\textbf{Challenge.} Evaluating the popularity of online comments is more challenging than assessing intrinsic text quality alone. In social media, popularity is not a fixed property of the comment itself, but a conditional outcome jointly shaped by the comment, the source content, the platform context, and the exposed audience:
\begin{equation}
    \pi(c \mid x, p, a)
\end{equation}
where \(c\) denotes the comment, \(x\) the source content, \(p\) the platform context, and \(a\) the audience state. This formulation makes explicit that popularity depends on both contextual conditions and audience structure.

However, \textbf{existing benchmarks} and evaluation protocols mainly focus on semantic similarity, fluency, or isolated stylistic aspects, with limited ability to capture how these factors interact with real engagement mechanisms. A single popularity predictor reduces dissemination to a \textbf{coarse platform-level estimate}, while \textbf{generic judge-style evaluation} ignores exposure conditions, audience heterogeneity, and variation in user responses. Recent in-the-wild evidence from public human--LLM interactions on social media further shows that engagement in multi-party environments is highly asymmetric and socially embedded, making popularity inseparable from exposure context and audience structure \cite{migliarini2026grokset}. As a result, current evaluation settings remain insufficient to determine whether a generated comment truly has real-world popularity potential.

\textbf{Motivation and Design Rationale.} These limitations motivate a benchmark design that approximates popularity from multiple complementary perspectives:
\begin{itemize}[leftmargin=*]
    \item \textbf{Content Quality} evaluates whether a generated comment is semantically appropriate and stylistically expressive, providing the textual foundation of popularity.
    \item \textbf{Popularity Prediction} captures platform-level interaction trends learned from real-world engagement data, reflecting aggregate platform tendencies.
    \item \textbf{User Behavior Simulation} models heterogeneous audience responses at the user level. Rather than acting as a generic LLM judge, it serves as a \textbf{data-grounded audience modeling component}. User profiles and agent weights are allocated according to real-world demographic statistics, regional internet population characteristics, and platform-specific user category distributions, ensuring that simulated reactions are conditioned on realistic audience composition.
\end{itemize}
Together, these three dimensions provide a structured approximation of popularity under heterogeneous exposure conditions, avoiding the reduction of complex popularity mechanisms to a single score and yielding a more faithful evaluation framework for socially grounded comment generation.

\begin{figure*}[t]
    \centering
    \includegraphics[width=1\linewidth]{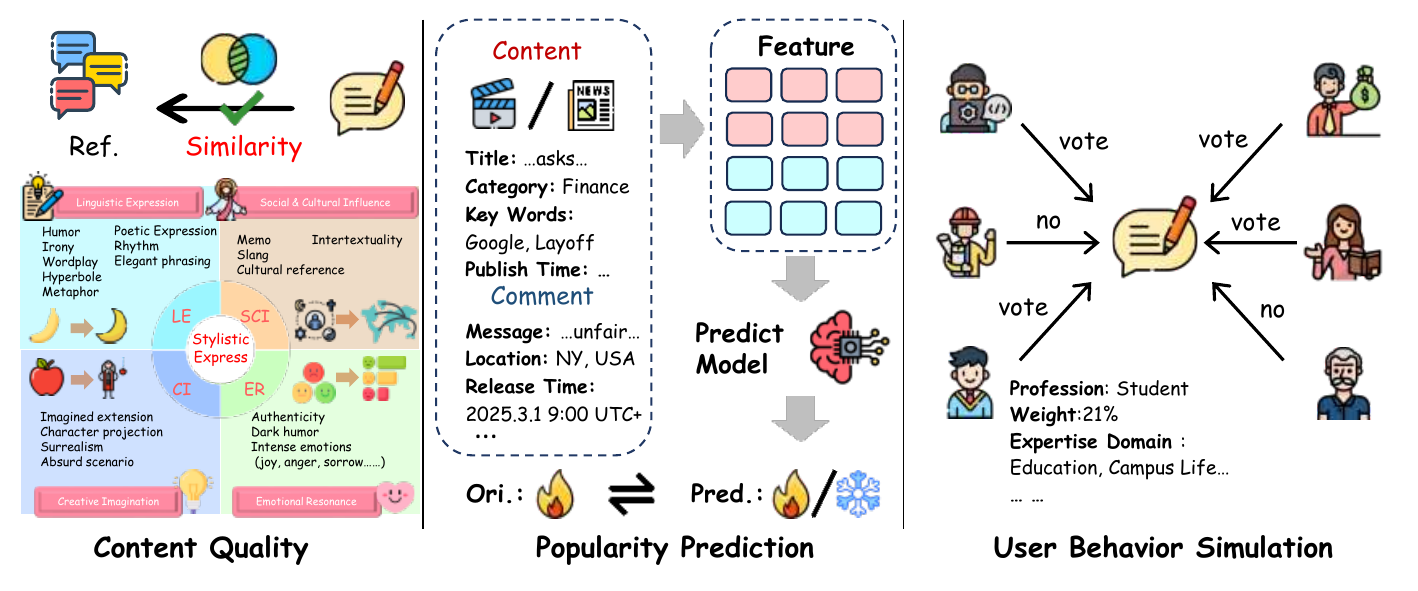}
    \caption{\textbf{HotComment} evaluates models from three key aspects:(1) \textbf{Content Quality}, assessed through multi-dimensional semantic comparison with top-\textit{k} popular comments in the dataset; (2) \textbf{Popularity Prediction}, based on trends from popularity prediction models trained on real-world interaction data; (3) \textbf{User Behavior Simulation}, conducted via agent-based modeling of online user behavior such as thumbs-up.}
    \label{fig:benchmark}
\end{figure*}

\section{HotComment Benchmark}
\label{sec: HotComment}

We first compare HotComment with the previous benchmark in Tab.\ref{tab:benchmark_comparison}.
Then, we introduce the three core dimensions of \textbf{HotComment}: 
\textbf{Content Quality}, which assesses linguistic and stylistic express; 
\textbf{Popularity Prediction}, which estimates engagement likelihood based on real-world interaction data; 
and \textbf{User Behavior Simulation}, which models audience preferences and interaction patterns to reflect realistic social dynamics.

\subsection{Dataset Construction}
\paragraph{Data Sources}
We construct the \textbf{HotComment} dataset by collecting large-scale article–comment and video–comment pairs from multiple mainstream online platforms, including NetEase News, Tencent News, and Bilibili.
These sources cover both textual and audiovisual content, enabling the study of multimodal communicative behavior and stylistic variation across platforms.
\paragraph{Scale and Composition}
The dataset comprises \textbf{over 43,000 online articles} and \textbf{34,000 videos}, encompassing approximately \textbf{1.4 million content-comment pairs} in total.
Each article or video is paired with corresponding user comments that reflect diverse linguistic styles and engagement behaviors.
For each item, we collect both \textbf{popular comments} with high user engagement and \textbf{non-popular comments} with relatively lower engagement; the average number of comments per item varies by content type and platform characteristics.

\paragraph{Popularity Labeling}
High-quality (popular) comments are defined as those ranked within the \textbf{top-15 by likes}, with like counts exceeding either \textbf{10\% of the top comment} or an absolute threshold of \textbf{2,000 likes}.
For low-quality (non-popular) comments, we select those posted on the \emph{same day} as the content with like counts not exceeding \textbf{10}, collecting at most 10 such comments per item.
This hybrid criterion ensures balanced representation while mitigating temporal bias.

\subsection{Evaluation Methods}

\subsubsection{Content Quality}
\label{subsec: content quality}
To comprehensively assess the intrinsic quality of generated comments, we evaluate not only their \textbf{semantic similarity} to human-written references but also their \textbf{stylistic expressiveness}\cite{lei-etal-2025-godbench} across four complementary dimensions:

\begin{itemize}[leftmargin=*]
    \item \textbf{Linguistic Expression} evaluates the rhetorical and writing artistry of comments, focusing on linguistic creativity such as humor, irony, metaphor, rhythm, and aesthetic fluency that enhance expressiveness and readability.
    
    \item \textbf{Creative Imagination} measures the degree of originality and associative thinking within comments, capturing the ability to connect distant concepts or construct unexpected, imaginative scenarios that extend semantic boundaries.
    
    \item \textbf{Emotional Resonance} examines the emotional depth and attitudinal stance of a comment, emphasizing its capacity to evoke empathy, convey genuine sentiment, or express strong affective tones that engage readers.
    
    \item \textbf{Social and Cultural Influence} assesses the comment’s potential for social propagation, including the use of memes, cultural references, and intertextual expressions that facilitate sharing, imitation, and collective resonance across communities.
\end{itemize}


\subsubsection{Popularity Prediction}
This dimension models \textbf{platform-level engagement preferences} to estimate how likely a comment would become popular within a given social media environment. 
Different platforms exhibit distinct user cultures and interaction mechanisms, leading to diverse definitions of ``popularity''. 
To reflect these differences, we train an individual \textbf{prediction model} for each platform using a fine-tuned BERT encoder with a task-specific MLP head, with real interaction data converted into binary popularity labels as supervision signals. 
Each model jointly encodes the contextual information of the article or video and the associated comment, and outputs a predicted popularity score that reflects engagement likelihood.
Recent benchmark construction for social-media popularity prediction has also begun to emphasize temporal alignment and temporal dynamics, suggesting that engagement modeling should consider not only content--comment matching but also time-sensitive propagation patterns \cite{xu2025smtpd}. 

Training is performed under a binary classification objective, where popular and non-popular comments serve as positive and negative samples, respectively. 
We employ a combination of \textit{cross-entropy loss} to ensure accurate classification and a \textit{supervised contrastive loss} to enhance representation discrimination between high- and low-engagement comments. 
This setup enables the model to learn platform-specific engagement patterns while providing a reliable data-driven metric for evaluating the real-world popularity potential of generated comments.
This design is broadly consistent with prior studies on popularity prediction and diffusion modeling, which treat engagement as a function of structural spread patterns, interaction dynamics, or jointly encoded content signals \cite{cheng2014can, goel2016structural, cao2020popularity}.

\subsubsection{User Behavior Simulation}
\textbf{User Behavior Simulation} models the composition of the exposed audience to approximate how comments would be perceived by different user groups within a realistic social environment. 
We formulate the exposure process as a \textbf{two-level hierarchical simulation}. 

\begin{itemize}[leftmargin=*]
    \item At the top level, we classify potential viewers into two primary categories: \textbf{interested users} and \textbf{casual viewers}. For each platform--domain pair, we first assign an \textbf{Exposure Specificity Index (ESI)} that represents the baseline exclusivity of audience exposure. A \textbf{domain-specialized agent} analyzes the content to determine audience exposure patterns based on five key determinants of selective exposure: \textbf{channel verticality, distribution channel characteristics, event salience, emotional arousal level,} and \textbf{celebrity or authority involvement}. The agent outputs an adjusted proportion, denoted as $p_I^*$, representing the estimated share of interested users in the total audience. This modeling approach aims to reproduce the realistic composition of audiences who \textit{actually see} the content, aligning with established theories of selective exposure, incidental news contact, and high-arousal dissemination.

    \item At the lower level, we derive the distribution of user subgroups from \textit{publicly accessible regional internet demographics} and \textit{platform-specific user category statistics}, rather than preset rules. These data act as \textit{data-driven priors} to construct heterogeneous audience segments with distinct interaction tendencies. This hierarchical design enables a user-centered interpretation of comment popularity beyond aggregate platform metrics. 
\end{itemize}

Through this hierarchical design, the simulation offers a population-level perspective on how comments are likely to be received, thereby complementing platform-level popularity prediction with \textbf{user-centric interpretability}.
Our formulation is also related to recent LLM-agent simulation frameworks, which emphasize role-conditioned behavior, social interaction, and aggregate engagement dynamics in synthetic but data-grounded online environments \cite{park2023generative, torberg2023simulating, li2024fine-grained, qiu2025llms}.
This perspective is further supported by recent multi-agent social-media simulation work, which models public-opinion evolution through cognitively grounded agents and dynamic interaction environments \cite{zhang2026posim}.
    
\begin{figure}[t]
    \centering
    \includegraphics[width=\linewidth]{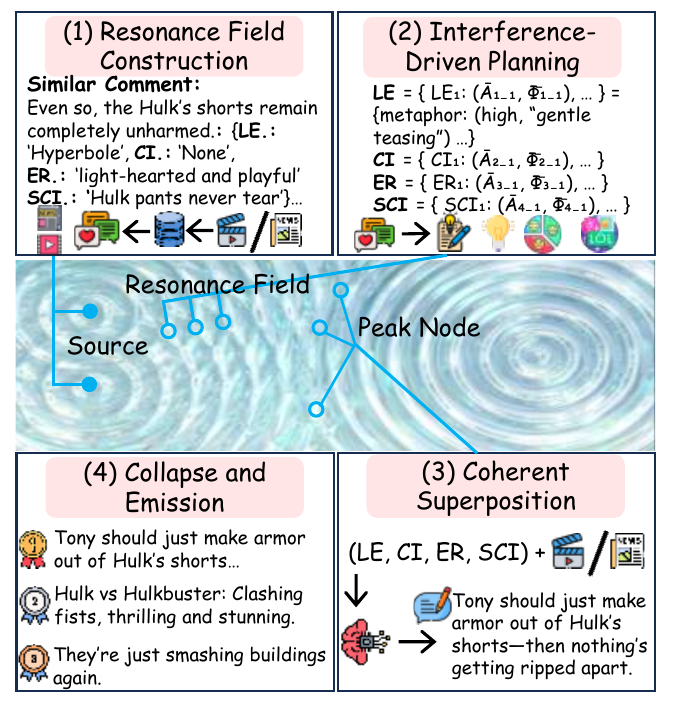}
    \vspace{-6pt}
    \caption{
        \textbf{Overview of the StyleCmt Framework.} 
        The comment generation process is modeled as a form of wave interference in a stylistic field. 
        The pipeline illustrates four consecutive stages:
        (1) \textit{Resonance Field Construction}. Retrieving similar hot comments and decomposing their stylistic components across four dimensions (LE, CI, ER, SCI);
        (2) \textit{Interference-Driven Planning}. Aggregating and identifying dominant stylistic patterns to form an interference blueprint;
        (3) \textit{Coherent Superposition}. Generating multiple linguistic realizations under the same stylistic configuration; and
        (4) \textit{Collapse and Emission} – selecting and refining the most resonant comment as the final output. 
    }
    \label{fig:framework2}
    \vspace{-8pt}
\end{figure}

\section{StyleCmt Framework}
\label{sec:StyleCmt}

\subsection{Motivation and Computational Grounding}
\label{subsec:motivation}

The \textbf{StyleCmt} framework is inspired by the way ideas and expressions spread through social interaction and gradually form shared preferences within a community. In our framework, historical comments under the same content context collectively define a \textbf{stylistic resonance field}, which captures the dominant expressive tendencies of the comment section.

To model this process, we represent each stylistic dimension as a vector with two components: \textbf{intensity}, which reflects the strength of community preference for that style, and \textbf{orientation}, which encodes its semantic and affective direction. When the stylistic representation of a generated comment is consistent with dominant historical patterns, the comment exhibits stronger stylistic coherence and better matches community preference. When the alignment is weak, the generated expression is less coherent with the surrounding discourse. This formulation provides a principled way to model stylistic preference in vector space and supports the generation of comments that are more consistent with collective audience expectations.

\subsection{Framework Overview and Modeling Process}
\label{subsec:framework_overview}

\paragraph{Resonance Field Construction.}
This step performs \textbf{contextual style distribution modeling}. To capture how a community tends to express itself, we construct the stylistic resonance field based on historical comments. Each comment is decomposed into four stylistic dimensions: Linguistic Expression, Creative Imagination, Emotional Resonance, and Social/Cultural Influence. For each dimension $i$, we estimate two properties from the data: (1) an \emph{intensity} scalar $\bar{A}_i(\mathbf{x})$, representing how strongly the community prefers this stylistic pattern at context $\mathbf{x}$, and (2) an \emph{orientation} unit vector $\bar{\mathbf{d}}_i(\mathbf{x})$, describing its typical stylistic inclination.
We summarize the overall contextual stylistic tendency as a weighted vector aggregation:
\begin{equation}
\mathbf{\Psi}_0(\mathbf{x}) = \sum_{i=1}^{4} \bar{A}_i(\mathbf{x})\, \bar{\mathbf{d}}_i(\mathbf{x}),
\end{equation}
which acts as a compact mathematical representation of the community's baseline preference.

\paragraph{Interference-Driven Planning.}
This phase executes \textbf{alignment-driven style planning}. Given the base preference $\mathbf{\Psi}_0$, we determine how a newly generated comment should adjust its stylistic mix. A candidate comment introduces a set of controllable stylistic contributions:
\begin{equation}
\mathbf{v}_i = A_i\, \mathbf{d}_i,
\end{equation}
where $A_i$ denotes the intended intensity of stylistic pattern $i$, and $\mathbf{d}_i$ describes its stylistic direction.
To evaluate how well this candidate aligns with community tendencies, we compute the \emph{interaction score} via the dot product of their orientation vectors (equivalent to cosine similarity):
\begin{equation}
I_{ij} = \mathbf{d}_i \cdot \bar{\mathbf{d}}_j.
\end{equation}
A positive value indicates that the intended direction $\mathbf{d}_i$ is compatible with the community inclination $\bar{\mathbf{d}}_j$, conceptually mimicking constructive interference. The planning objective selects $\{A_i, \mathbf{d}_i\}$ that maximize the overall alignment:
\begin{equation}
\mathcal{E}_{\text{align}} = \sum_{i<j} A_i A_j I_{ij},
\end{equation}
which prefers stylistic combinations that mutually reinforce each other and fit the contextual distribution.

\paragraph{Coherent Superposition.}
Computationally, this stage represents \textbf{multi-dimensional feature aggregation and decoding}. Using the optimized style parameters, the generator produces multiple candidate comments by sampling from the aggregated stylistic space. Each candidate is represented in the latent space as:
\begin{equation}
\mathbf{\Psi}_k = \mathbf{\Psi}_0 + \sum_{i=1}^{4} A_{k,i}^* \mathbf{d}_{k,i}^* + \mathbf{\epsilon}_k,
\end{equation}
where $A_{k,i}^*$ and $\mathbf{d}_{k,i}^*$ reflect the actual stylistic features realized during text decoding, and $\mathbf{\epsilon}_k$ denotes minor sampling variations. 
To evaluate the coherence of each candidate with respect to community preference, we compute a standard cosine similarity score between the realized representation and the baseline preference:
\begin{equation}
C_k = \frac{\mathbf{\Psi}_k \cdot \mathbf{\Psi}_0}{\|\mathbf{\Psi}_k\|\,\|\mathbf{\Psi}_0\|}.
\end{equation}
Higher scores indicate that the generated candidate naturally resonates with the historical community tendencies.

\paragraph{Collapse and Emission.}
Finally, acting as a \textbf{resonance-aware candidate selection} mechanism, this stage finalizes the text output. In probabilistic text generation, a model's continuous probability distribution effectively "collapses" into a single discrete sequence. We select the candidate with the highest coherence:
\begin{equation}
\mathbf{\Psi}^* = \arg\max_k\, C_k.
\end{equation}
The selected comment undergoes minimal refinement for clarity and safety. The resulting output represents the expression that best aligns with the community’s stylistic preferences, successfully materializing the computed social resonance into readable text.

\definecolor{darkred}{RGB}{255,102,102}
\definecolor{midred}{RGB}{255, 153, 153}
\definecolor{lightred}{RGB}{255, 204,204}

\definecolor{lightpurple}{RGB}{216,191,216}  
\definecolor{lightorange}{RGB}{255, 204, 153} 
\definecolor{lightgray}{RGB}{224,224,224}     

\newcommand{\first}[1]{\mbox{\colorbox{lightpurple}{\strut #1}}}
\newcommand{\second}[1]{\mbox{\colorbox{lightorange}{\strut #1}}}
\newcommand{\third}[1]{\mbox{\colorbox{lightgray}{\strut #1}}}

\newcommand{\twolines}[2]{%
  \begin{tabular}{@{}c@{}}%
    #1\\[-8.0pt]%
    {\scriptsize (#2)}%
  \end{tabular}%
}

\newcommand{\heat}[3]{%
  \begingroup
  \pgfmathsetmacro{\clamped}{max(0,#1)}%
  \pgfmathsetmacro{\denom}{max(#3,0)}%
  \pgfmathsetmacro{\ratio}{ifthenelse(\denom>0, 100*\clamped/\denom, 0)}%
  \pgfmathparse{round(\ratio)}%
  \xdef\mix{\pgfmathresult}%
  \cellcolor{lightpurple!\mix!white}{#2}%
  \endgroup
}

\newcommand{\colboxwidth}{1.0cm}

\newcommand{\twolinesheat}[4]{%
  \begingroup
  \pgfmathsetmacro{\clamped}{max(0,#3)}%
  \pgfmathsetmacro{\denom}{max(#4,0)}%
  \pgfmathsetmacro{\ratio}{ifthenelse(\denom>0, 100*\clamped/\denom, 0)}%
  \pgfmathparse{round(\ratio)}\xdef\mix{\pgfmathresult}%
  \makebox[\colboxwidth][c]{%
    \colorbox{lightpurple!\mix!white}{%
      \parbox[c]{\colboxwidth}{\centering
        \strut #1\\[-2pt]
        {\scriptsize #2}\strut
      }%
    }%
  }%
  \endgroup
}

    \begin{table}[t]
    \small
    \centering
    \renewcommand{\arraystretch}{1.2}
    \setlength{\tabcolsep}{2pt}
    
    \resizebox{\columnwidth}{!}{
    \begin{tabular}{@{}l c *{4}{c} c c@{}}
    \toprule
    \multirow{2}{*}{\textbf{Models}} & 
    \multicolumn{4}{c}{\textbf{Content Quality}} & 
    \textbf{Popularity} & 
    \multirow{2}{*}{\textbf{UBS}} \\
    \cmidrule(lr){2-5}
    & BLEU-1 & METEOR & F1 & SRS & \textbf{Prediction} &  \\  
    \midrule
    Mistral-7B & 8.11 & 9.75 & 53.89 & 49.58 & 69.63 & 63.71 \\  
    Baichuan2-7B & 7.34 & 8.94 & 57.04 & 39.58 & 57.39 & 31.68 \\
    \midrule
    Qwen2.5-0.5B  & 14.22 & 13.41 & 57.79 & 31.34 & 66.69 & 62.23 \\
    Qwen2.5-7B   & 17.08 & 17.14 & 58.63 & 45.17 & 75.84 & \third{76.71} \\
    Qwen2.5-14B  & 15.36 & 15.64 & 58.02 & 51.78 & \third{76.70} & 70.19 \\
    \quad +\textbf{StyleCmt} & \first{21.08} & \third{18.61} & \second{63.18} & \first{60.78} & \first{82.18} & \first{84.50} \\
    \midrule
    LLaMA3.1-8B   & 17.03 & 16.64 & 57.78 & 47.66 & 47.25 & 63.99 \\
    \quad +\textbf{StyleCmt} & \third{20.00} & \second{20.48} & \third{60.90} & \third{57.74} & 73.52 & 71.49 \\
    \midrule
    ChatGPT-4o & 16.97 & 16.41 & 59.78 & 57.09 & 71.41 & 72.38 \\
    \quad +\textbf{StyleCmt} & \second{20.57} & \first{21.15} & \first{63.98} & \second{59.49} & \second{76.98} & \second{77.69} \\
    \bottomrule
    \end{tabular}
    }
    \vspace{0.5em}
    \caption{
    \textbf{Results of text-based comment generation using large language models (LLMs).}
    This table reports quantitative results of baseline and enhanced models on the HotComment benchmark.  
    \textbf{F1} denotes the BERTScore F1 metric.  
    \textbf{SRS} is the averaged stylistic quality score derived from four stylistic dimensions (\textit{LE, CI, ER, SCI}) under \textbf{Content Quality}.  
    \textbf{UBS} represents the User Behavior Simulation score reflecting user-level response patterns.  
    Cells highlighted in color indicate the top-three results within each column: 
    \mbox{\colorbox{lightpurple}{\strut 1st}}, 
    \mbox{\colorbox{lightorange}{\strut 2nd}}, and 
    \mbox{\colorbox{lightgray}{\strut 3rd}}.
    }
    \label{tab:hotcommentbench_results}
    \end{table}

\begin{table}[t]
\small
\centering
\renewcommand{\arraystretch}{1.2}
\setlength{\tabcolsep}{2pt}

\resizebox{\columnwidth}{!}{
\begin{tabular}{@{}l c *{4}{c} c c@{}}
\toprule
\multirow{2}{*}{\textbf{Models}} & 
\multicolumn{4}{c}{\textbf{Content Quality}} & 
\textbf{Popularity} & 
\multirow{2}{*}{\textbf{UBS}} \\
\cmidrule(lr){2-5}
& BLEU-1 & METEOR & F1 & SRS & \textbf{Prediction} &  \\  
\midrule

Mistral-3.1-24B & \third{15.32} & \third{16.07} & 60.68 & 54.25 & \third{74.22} & \second{76.08} \\
Gemini2.5-Image& 15.25 & 13.69 & \third{61.34} & \third{56.63} & 72.05 & 73.09 \\
ChatGPT-4o & 14.84 & 12.65 & 61.21 & 55.71 & 72.06 & 73.01 \\
ChatGPT-4o-mini & 13.25 & 10.47 & 61.21 & 52.56 & 70.03 & 71.07 \\
\midrule

Qwen3-VL-4B & 12.75 & 13.31 & 58.42 & 34.38 & 68.03 & 64.22 \\
Qwen3-VL-8B & 14.55 & 15.59 & 59.03 & 49.81 & 72.52 & 70.58 \\
\quad +\textbf{StyleCmt}& \first{19.27} & \first{19.53} & \first{62.78} & \first{60.98} & \first{82.06} & \first{84.09} \\
\midrule

LLaMA3.2-Vision & 13.82 & 13.84 & 60.04 & 49.43 & 62.03 & 65.08 \\
\quad +\textbf{StyleCmt}& \second{17.58} & \second{16.89} & \second{62.57} & \second{57.42} & \second{75.07} & \third{75.04} \\

\bottomrule
\end{tabular}
}
\vspace{0.5em}
\caption{
\textbf{Results of multimodal comment generation using vision-language models (MLLMs).}
\textbf{F1} denotes the BERTScore F1 metric.
\textbf{SRS} is the averaged stylistic quality score derived from four stylistic dimensions (\textit{LE, CI, ER, SCI}) under \textbf{Content Quality}.
\textbf{UBS} represents the User Behavior Simulation score reflecting user-level response patterns.  
Cells highlighted in color indicate the top-three results within each column: 
\mbox{\colorbox{lightpurple}{\strut 1st}}, 
\mbox{\colorbox{lightorange}{\strut 2nd}}, and 
\mbox{\colorbox{lightgray}{\strut 3rd}}.
}
\label{tab:hotcommentbench_mllm_results}
\end{table}

\section{Experiments}
\label{sec: experiment}
    
\subsection{Evaluation Metrics}
We evaluate model performance along three dimensions: \textbf{content quality}, \textbf{popularity prediction}, and \textbf{user behavior simulation}.

\textbf{Content Quality.}
To assess semantic adequacy under the open-ended, multi-reference nature of comment generation, we compute similarity exclusively against \textbf{popular comments} from the dataset.
We follow a weighted multi-reference strategy inspired by W-BLEU\cite{qin2018automatic}, where each reference comment is assigned a weight determined by its real-world engagement level.
Engagement values are mapped onto a Gaussian distribution constrained to 
[0.6,1.0], giving more influential references a higher contribution while preserving diversity among less salient ones.
The final score for each generated comment is the weighted maximum similarity across BLEU-1, METEOR, and BERTScore F1
For \textbf{stylistic evaluation}, we employ both \textit{Qwen3-14B} (using a publicly released checkpoint) and \textit{GPT-4o} as independent expert evaluators.
Scores from the two models are averaged to reduce evaluator bias and avoid collapsing stylistic judgments onto the preference of any single evaluator.
We report dimension-wise scores on Linguistic Expression, Creative Imagination, Emotional Resonance, and Social/Cultural Influence, and their mean forms the \textbf{Stylistic Resonance Score (SRS)}.

\textbf{Popularity Prediction.} To measure alignment with real engagement patterns, semantic features from the generated comment are combined with \textbf{textual metadata} from the source content (title, keywords, and description). A trained prediction model outputs a normalized popularity score that reflects expected audience interaction. The predictor itself is reliable, achieving an accuracy of 82.61 and an F1 score of 79.39; details and full results are provided in the Appendix.

\textbf{User Behavior Simulation.} To approximate user reactions, we perform agent-based simulation using Qwen3-14B. Given a piece of content and a generated comment, the simulator estimates an \textbf{engagement score} reflecting the likelihood of user interaction, offering a behavioral view of comment effectiveness. As additional evidence of validity, on a within-item ranking test the simulator achieves a mean NDCG of 70.13 with Qwen3-14B and 68.34 with ChatGPT-4o as evaluators; full experimental details are deferred to the Appendix.

\subsection{Experimental Setups}

\paragraph{Dataset Partitioning.} 
To ensure a reliable and unbiased evaluation, the dataset is divided into \textbf{training}, \textbf{validation}, and \textbf{test} sets following an 8:1:1 ratio. 
To prevent temporal and topical leakage, samples are stratified according to both \textbf{publication time} and \textbf{content category}, ensuring that later-published content does not share topics with earlier training data. 
This temporal–categorical balance effectively mitigates overfitting caused by event recency and maintains domain diversity across splits.

\paragraph{Implementation Details.} 
The \textit{Popularity Prediction} model in our benchmark is trained on the training split, while evaluation is performed exclusively on the held-out test set. 
To ensure fairness, all open-source models are initialized from their official instruction-tuned checkpoints, and no additional fine-tuning or task adaptation is applied during evaluation.
Experiments are conducted on \textbf{NVIDIA A100 GPUs}. All open-source models are deployed with \textbf{int8 quantization} to enable efficient inference within GPU memory constraints.
For closed-source models such as GPT-4o, we access them through standardized API interfaces to maintain consistent prompt formatting and evaluation settings across modalities.

\begin{table}[t]
\small
\centering
\renewcommand{\arraystretch}{1.35}
\setlength{\tabcolsep}{2pt}

\resizebox{\columnwidth}{!}{
\begin{tabular}{@{}l *{4}{c} c c@{}}
\toprule
\multirow{2}{*}{\textbf{Models}} & 
\multicolumn{4}{c}{\textbf{Content Quality}} & 
\textbf{Popularity} & 
\multirow{2}{*}{\textbf{UBS}} \\
\cmidrule(lr){2-5}
& BLEU-1 & METEOR & F1 & SRS & \textbf{Prediction} &  \\  
\midrule

Qwen2.5-14B 
& 15.36 & 15.64 & 58.02 & 51.78 & 76.70 & 70.19 \\

\quad +CoT 
& \twolinesheat{17.01}{(+10.74\%)}{10.74}{37.24}
& \twolinesheat{16.64}{(+6.39\%)}{6.39}{37.24}
& \twolinesheat{59.02}{(+1.72\%)}{1.72}{37.24}
& \twolinesheat{54.35}{(+4.96\%)}{4.96}{37.24}
& \twolinesheat{78.95}{(+2.93\%)}{2.93}{37.24}
& \twolinesheat{79.82}{(+13.72\%)}{13.72}{37.24} \\

\quad +5-shot 
& \twolinesheat{17.80}{(+15.89\%)}{15.89}{37.24}
& \twolinesheat{16.27}{(+4.03\%)}{4.03}{37.24}
& \twolinesheat{60.92}{(+5.00\%)}{5.00}{37.24}
& \twolinesheat{55.72}{(+7.61\%)}{7.61}{37.24}
& \twolinesheat{77.37}{(+0.87\%)}{0.87}{37.24}
& \twolinesheat{76.77}{(+9.37\%)}{9.37}{37.24} \\

\quad +\textbf{StyleCmt (Ours)} 
& \twolinesheat{21.08}{(+37.24\%)}{37.24}{37.24}
& \twolinesheat{18.61}{(+18.99\%)}{18.99}{37.24}
& \twolinesheat{63.18}{(+8.89\%)}{8.89}{37.24}
& \twolinesheat{60.78}{(+17.38\%)}{17.38}{37.24}
& \twolinesheat{82.18}{(+7.14\%)}{7.14}{37.24}
& \twolinesheat{84.50}{(+20.39\%)}{20.39}{37.24} \\

\midrule
LLaMA3.1-8B
& 17.03 & 16.64 & 57.78 & 47.66 & 47.25 & 63.99 \\

\quad +CoT 
& \twolinesheat{17.09}{(+0.35\%)}{0.35}{55.60}
& \twolinesheat{16.53}{(-0.66\%)}{-0.66}{55.60}
& \twolinesheat{60.38}{(+4.50\%)}{4.50}{55.60}
& \twolinesheat{47.83}{(+0.36\%)}{0.36}{55.60}
& \twolinesheat{59.79}{(+26.54\%)}{26.54}{55.60}
& \twolinesheat{67.99}{(+6.25\%)}{6.25}{55.60} \\

\quad +5-shot 
& \twolinesheat{18.55}{(+8.93\%)}{8.93}{55.60}
& \twolinesheat{17.58}{(+5.65\%)}{5.65}{55.60}
& \twolinesheat{60.31}{(+4.38\%)}{4.38}{55.60}
& \twolinesheat{52.89}{(+10.97\%)}{10.97}{55.60}
& \twolinesheat{67.54}{(+42.94\%)}{42.94}{55.60}
& \twolinesheat{70.33}{(+9.91\%)}{9.91}{55.60} \\

\quad +\textbf{StyleCmt (Ours)} 
& \twolinesheat{20.00}{(+17.44\%)}{17.44}{55.60}
& \twolinesheat{20.48}{(+23.08\%)}{23.08}{55.60}
& \twolinesheat{60.90}{(+5.40\%)}{5.40}{55.60}
& \twolinesheat{57.74}{(+21.15\%)}{21.15}{55.60}
& \twolinesheat{73.52}{(+55.60\%)}{55.60}{55.60}
& \twolinesheat{71.49}{(+11.72\%)}{11.72}{55.60} \\

\bottomrule
\end{tabular}
}
\vspace{0.5em}
\caption{
\textbf{Comparison of StyleCmt with reasoning-based methods on text-based comment generation (LLMs).}
Each enhancement row reports the absolute score (top) and relative improvement (bottom, in parentheses).
\textbf{SRS} denotes the averaged stylistic resonance score, and \textbf{UBS} represents simulated user engagement.
A continuous color gradient is applied within each model group to indicate the magnitude of performance gain, where deeper shading reflects greater relative improvement.
}
\label{tab:llm_compare_two_line}
\end{table}

\begin{table}[t]
\small
\centering
\renewcommand{\arraystretch}{1.35}
\setlength{\tabcolsep}{2pt}

\resizebox{\columnwidth}{!}{
\begin{tabular}{@{}l *{4}{c} c c@{}}
\toprule
\multirow{2}{*}{\textbf{Models}} & 
\multicolumn{4}{c}{\textbf{Content Quality}} & 
\textbf{Popularity} & 
\multirow{2}{*}{\textbf{UBS}} \\
\cmidrule(lr){2-5}
& BLEU-1 & METEOR & F1 & SRS & \textbf{Prediction} &  \\  
\midrule

Qwen3-VL-8B
& 14.55 & 15.59 & 59.03 & 49.81 & 72.52 & 70.58 \\

\quad +CoT 
& \twolinesheat{15.46}{(+6.25\%)}{6.25}{32.44}
& \twolinesheat{16.17}{(+3.72\%)}{3.72}{32.44}
& \twolinesheat{60.93}{(+3.22\%)}{3.22}{32.44}
& \twolinesheat{52.88}{(+6.16\%)}{6.16}{32.44}
& \twolinesheat{78.27}{(+7.93\%)}{7.93}{32.44}
& \twolinesheat{75.54}{(+7.03\%)}{7.03}{32.44} \\

\quad +5-shot 
& \twolinesheat{16.52}{(+13.54\%)}{13.54}{32.44}
& \twolinesheat{16.81}{(+7.83\%)}{7.83}{32.44}
& \twolinesheat{61.59}{(+4.34\%)}{4.34}{32.44}
& \twolinesheat{54.79}{(+10.00\%)}{10.00}{32.44}
& \twolinesheat{78.87}{(+8.76\%)}{8.76}{32.44}
& \twolinesheat{76.25}{(+8.03\%)}{8.03}{32.44} \\

\quad +\textbf{StyleCmt (Ours)} 
& \twolinesheat{19.27}{(+32.44\%)}{32.44}{32.44}
& \twolinesheat{19.53}{(+25.27\%)}{25.27}{32.44}
& \twolinesheat{62.78}{(+6.35\%)}{7.71}{32.44}
& \twolinesheat{60.98}{(+22.43\%)}{22.43}{32.44}
& \twolinesheat{82.06}{(+13.15\%)}{13.15}{32.44}
& \twolinesheat{84.09}{(+19.14\%)}{19.14}{32.44} \\

\midrule

LLaMA3.2-Vision-11B
& 13.82 & 13.84 & 60.04 & 49.43 & 62.03 & 65.08 \\

\quad +CoT 
& \twolinesheat{14.68}{(+6.22\%)}{6.22}{27.21}
& \twolinesheat{14.29}{(+3.25\%)}{3.25}{27.21}
& \twolinesheat{61.83}{(+2.98\%)}{2.98}{27.21}
& \twolinesheat{50.98}{(+3.14\%)}{3.14}{27.21}
& \twolinesheat{65.04}{(+4.85\%)}{4.85}{27.21}
& \twolinesheat{70.07}{(+7.67\%)}{7.67}{27.21} \\

\quad +5-shot 
& \twolinesheat{15.62}{(+13.02\%)}{13.02}{27.21}
& \twolinesheat{14.87}{(+7.44\%)}{7.44}{27.21}
& \twolinesheat{61.89}{(+3.08\%)}{3.08}{27.21}
& \twolinesheat{54.44}{(+10.14\%)}{10.14}{27.21}
& \twolinesheat{70.01}{(+12.86\%)}{12.86}{27.21}
& \twolinesheat{72.53}{(+11.45\%)}{11.45}{27.21} \\

\quad +\textbf{StyleCmt (Ours)} 
& \twolinesheat{17.58}{(+27.21\%)}{27.21}{27.21}
& \twolinesheat{16.89}{(+22.04\%)}{22.04}{27.21}
& \twolinesheat{62.57}{(+4.21\%)}{5.88}{27.21}
& \twolinesheat{57.42}{(+16.16\%)}{16.16}{27.21}
& \twolinesheat{75.07}{(+21.02\%)}{21.02}{27.21}
& \twolinesheat{75.04}{(+15.30\%)}{15.30}{27.21} \\

\bottomrule

\end{tabular}
}
\vspace{0.5em}
\caption{
\textbf{Comparison of StyleCmt with reasoning-based methods on video-based comment generation (MLLMs).}
Each enhancement row reports the absolute score (top) and relative improvement (bottom, in parentheses).
\textbf{SRS} denotes the averaged stylistic resonance score, and \textbf{UBS} represents simulated user engagement.
A continuous color gradient is applied within each model group to indicate the magnitude of performance gain, where deeper shading reflects greater relative improvement.
}
\label{tab:ablation_mllm_results_twoline}
\end{table}

\subsection{Benchmark Results}

\paragraph{Text-based Comment Generation (LLMs).}
We evaluate a series of large language models, including \textit{Qwen2.5}\cite{qwen2025qwen25technicalreport}, \textit{LLaMA3.1}\cite{grattafiori2024llama3herdmodels}, \textit{ChatGPT-4o}\cite{openai2023gpt4}, \textit{Mistral}\cite{jiang2023mistral7b}, and \textit{Baichuan2}\cite{yang2025baichuan2openlargescale}, on the text-based comment generation task. 
Tab.~\ref{tab:hotcommentbench_results} shows that StyleCmt improves text-based comment generation across all evaluated large language models.  
Baseline models already demonstrate strong semantic quality, yet they often produce comments that lack stylistic alignment or social relevance.  
With StyleCmt applied, all LLMs exhibit consistent gains in both semantic similarity metrics and stylistic quality measures.  
Models such as Qwen2.5 and LLaMA3.1 show clear increases in BLEU-1, METEOR, F1, and SRS, while engagement-oriented metrics including Popularity Prediction and UBS also rise.  
The improvements are stable across model sizes and architectures, indicating that StyleCmt provides a generalizable enhancement to text-based comment generation.

\paragraph{Multimodal Comment Generation (MLLMs).}
We further evaluate multimodal large language models, including \textit{Qwen3-VL}\cite{qwen2025qwen25technicalreport}, \textit{LLaMA3.2-Vision}\cite{lee2025efficientllama32visiontrimmingcrossattended}, \textit{Mistral-3.1}\cite{jiang2023mistral7b}, \textit{Gemini2.5-Image}\cite{comanici2025gemini}, \textit{ChatGPT-4o}, and \textit{ChatGPT-4o-mini}\cite{openai2023gpt4}, on video–comment generation under the same test configuration. 
A similar trend appears in video-based generation.  
As reported in Tab.~\ref{tab:hotcommentbench_mllm_results}, multimodal models benefit from StyleCmt with consistent improvements in semantic accuracy, stylistic coherence, and engagement-related scores.  
Qwen3-VL and LLaMA3.2-Vision both show sizeable gains across BLEU-1, METEOR, SRS, and UBS, demonstrating that stylistic conditioning enhances multimodal grounding as well.  
Notably, open-source models equipped with StyleCmt achieve performance comparable to or exceeding certain closed-source baselines in several engagement metrics, suggesting that audience-aware stylistic modeling can compensate for differences in model scale.

\subsection{Performance of StyleCmt}

\paragraph{Comprehensive Comparison across Models.}
Across both LLMs and MLLMs, StyleCmt provides consistent improvements over all evaluated baselines.  
The gains extend across semantic similarity, stylistic quality, and engagement-focused metrics.  
Larger instruction-tuned models tend to benefit more from StyleCmt, although smaller models also exhibit notable increases.  
The overall pattern indicates that the proposed framework enhances linguistic expressiveness and social alignment in a model-agnostic manner, improving both content quality and predicted audience response.

\paragraph{Comparison with Other Strategies.}
We compare StyleCmt with two commonly used enhancement strategies: chain-of-thought prompting and few-shot prompting.  
Table\ref{tab:llm_compare_two_line} and \ref{tab:ablation_mllm_results_twoline} show that while these strategies provide moderate gains in several metrics, their improvements are generally limited and inconsistent.  
Chain-of-thought tends to increase semantic coherence, and the five-shot setting offers slightly larger gains in some cases, yet both methods produce only small changes in stylistic resonance and user engagement, with increases in UBS remaining within a narrow range.

In contrast, StyleCmt consistently produces larger and more stable improvements across all evaluated models.  
It yields substantial gains in BLEU-1, METEOR, SRS, Popularity Prediction, and UBS, often exceeding the increases achieved by chain-of-thought or few-shot prompting by a wide margin.  
A similar pattern appears in LLaMA3.1 and Qwen3-VL, where StyleCmt provides stronger improvements across content quality and stylistic coherence.  
Moreover, StyleCmt enhances engagement-related metrics more effectively, indicating that the generated comments align more closely with expressive and affective tendencies commonly observed in real discussion environments.

These findings show that conventional enhancement strategies primarily improve local semantic refinement but are limited in their ability to capture stylistic preference patterns or broader discourse resonance.  
StyleCmt directly models such preferences and their interaction with social context, resulting in comments that are more expressive, contextually appropriate, and more effective at eliciting engagement.  
This demonstrates that stylistic conditioning plays a central role in improving the communicative impact of generated comments in social media settings.

\begin{figure}[t]
    \centering
    \includegraphics[width=\columnwidth]{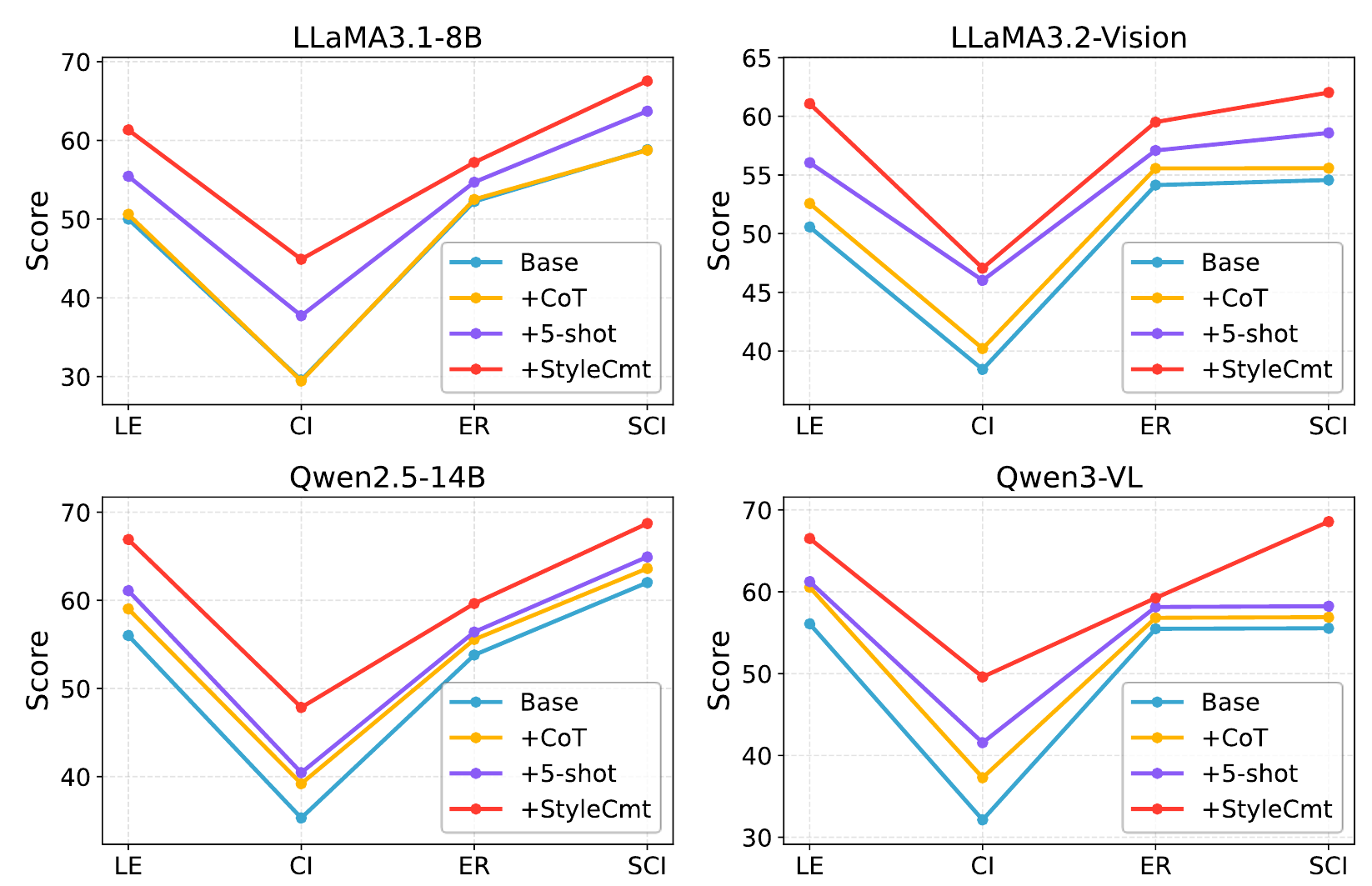}
    \caption{
    \textbf{Improvements across stylistic dimensions.} 
    Relative gains of StyleCmt on four dimensions of the \textbf{Stylistic Resonance Score (SRS)}: 
    \textbf{Linguistic Expression (LE)}, \textbf{Creative Imagination (CI)}, 
    \textbf{Emotional Resonance (ER)}, and \textbf{Social and Cultural Influence (SCI)}. 
    The upward trend demonstrates balanced enhancement of stylistic coherence and expressiveness.
    }
    \label{fig:line_srs}
\end{figure}

\paragraph{Stylistic Resonance Analysis.}  
To further examine the stylistic impact of StyleCmt, we analyze the relative improvements across the four dimensions that constitute the \textbf{Stylistic Resonance Score (SRS)}: \textbf{Linguistic Expression (LE)}, \textbf{Creative Imagination (CI)}, \textbf{Emotional Resonance (ER)}, and \textbf{Social and Cultural Influence (SCI)}.  
Figure~\ref{fig:line_srs} illustrates the percentage increases achieved by StyleCmt on each dimension for representative models.  
The results show steady gains across all stylistic aspects, confirming that StyleCmt enhances stylistic expressiveness in a balanced and interpretable manner.  

The most substantial improvements are observed in \textit{Creative Imagination} and \textit{Social and Cultural Influence}, where StyleCmt strengthens associative creativity and contextual relevance to audience culture.  
\textit{Linguistic Expression} and \textit{Emotional Resonance} also exhibit clear upward trends, reflecting smoother rhetorical structure and more natural affective tone.  
The consistent growth across dimensions suggests that StyleCmt amplifies stylistic coherence rather than optimizing isolated features.  

Overall, these results demonstrate that StyleCmt effectively models constructive interaction among stylistic components, leading to coordinated enhancement across expressive, imaginative, emotional, and social dimensions.  
The observed upward trajectories in the line chart highlight that stylistic resonance contributes directly to more engaging and audience-aligned comment generation.

\section{Conclusion}
\label{sec: conclusion}

In this work, we introduced \textbf{HotComment}, a large-scale benchmark for evaluating online comment generation across both textual and visual modalities. 
The benchmark integrates three complementary dimensions, including \textbf{Content Quality}, \textbf{Popularity Prediction}, and \textbf{User Behavior Simulation}, to assess linguistic expressiveness and social engagement potential in a unified framework. 
We further proposed \textbf{StyleCmt}, a wave-interference-inspired framework that models stylistic interactions to generate comments aligned with audience preferences. 
Extensive experiments demonstrate that StyleCmt consistently enhances content quality, stylistic richness, and engagement alignment across both LLMs and MLLMs, establishing a solid foundation for studying socially resonant comment generation.


\bibliographystyle{ACM-Reference-Format}
\bibliography{main}


\end{document}